\documentclass[11pt]{article}
\usepackage{amscd,amssymb,stmaryrd}
\usepackage{amsmath}
\usepackage{graphicx}
\usepackage{subcaption}
\usepackage[utf8]{inputenc}
\usepackage[export]{adjustbox}
\usepackage{wrapfig}
\usepackage{amstext}
\usepackage{pstricks,pst-node,pst-plot,pst-coil}
\usepackage{amsthm}
\usepackage{amsmath}
\usepackage{color}
\usepackage{mathtools}
\usepackage{hyperref}
\usepackage[english]{babel}
\usepackage{booktabs}
\usepackage{mathrsfs}
\usepackage{comment}
\usepackage{float}
\usepackage[linesnumbered,ruled,vlined, ruled,vlined]{algorithm2e}

\newcommand{\off}{\text{off}}
\newcommand{\snap}{\text{snap}}
\theoremstyle{definition}

\theoremstyle{remark}

\title{Multi-agent Reinforcement Learning Accelerated MCMC on Multiscale Inversion Problem}
\author{Eric Chung\thanks{Department of Mathematics, The Chinese University of Hong Kong, Shatin, Hong Kong}\ , \quad
Yalchin Efendiev\footnote{Department of Mathematics, Texas A\&M University, College Station, TX 77843, USA \& North-Eastern Federal University, Yakutsk, Russia}\, \quad 
Wing Tat Leung\thanks{Department of Mathematics, University of California, Irvine, CA 92697, USA}\ , \quad Sai-Mang Pun\thanks{Department of Mathematics, Texas A\&M University, College Station, TX 77843, USA}\ , \quad Zecheng Zhang\thanks{Department of Mathematics, Texas A\&M University, College Station, TX 77843, USA}}
\begin{document}
\maketitle
\begin{abstract}
In this work, we propose a multi-agent actor-critic reinforcement learning (RL) algorithm to accelerate the multi-level Monte Carlo Markov Chain (MCMC) sampling algorithms. The policies (actors) of the agents are used to generate the proposal in the MCMC steps; and the critic, which is centralized, is in charge of estimating the long term reward. We verify our proposed algorithm by solving an inverse problem with multiple scales. 
There are several difficulties in the implementation of this problem by using traditional MCMC sampling. Firstly, the computation of the posterior distribution involves
evaluating the forward solver, which is very time consuming for a problem with heterogeneous. We hence propose to use the multi-level algorithm. More precisely, 
we use the generalized multiscale finite element method (GMsFEM) as the forward solver in evaluating a posterior distribution in the multi-level rejection procedure. 
Secondly, it is hard to find a function which can generate samplings which are meaningful. To solve this issue, we learn an RL policy as the proposal generator.
Our experiments show that the proposed method significantly improves the sampling process.\end{abstract}

\section{Introduction}

The method of Monte Carlo Markov Chain (MCMC) has been widely used in solving inverse problems \cite{tan2014adaptive, mondal2014bayesian, efendiev2006preconditioning, chung2017generalized}. There are two issues which limit the computation speed of the MCMC algorithm. The first to mention is the evaluation of the forward solver. 
To compute the posterior distributions, computationally demanding simulations are needed for evaluating acceptance probabilities driven by forward problems.
This issue can be partially solved by applying the multi-level algorithms proposed in \cite{efendiev2006preconditioning}. Instead of using one solver, the multi-level algorithms apply different solvers and perform the multi-level acceptances/rejections. If the coarser solver rejects the proposal, then a new proposal is proposed. Since coarse solvers are easy to be evaluated, the rejection process can be accelerated. 

Multi-level algorithms can naturally be incorporated with the multiscale inverse problems. This is due to the nature of the multiscale finite element methods \cite{chung2018non, chung2016adaptive, chung2018constraint, chetverushkin2020computational}. Multiscale basis functions defined on a coarse grid of the computational domain are calculated by solving some local cell problems. Then, we can select different basis functions and use them in coarse forward solvers. In this work, we will use the generalized multiscale finite element methods (GMsFEM) as a forward solver.

Another issue associated with the MCMC is the proposal generator. The earliest methods developed by Metropolis \cite{metropolis1953equation} use the random walk to generate the proposal. Many other methods \cite{hastings1970monte, craiu2009learn,gilks1994adaptive,laloy2012high,hoffman2014no,duane1987hybrid,roberts1998optimal,girolami2011riemann,calderhead2014general,brooks1998general,laloy2013efficient,brockwell2006parallel, zhang2020improving} including using kernel adaptation, Hamiltonian dynamics, Langevin dynamics, parallel marginalization, kernel coupling and multiple chain simulation, parallelization have been proposed to improve the efficiency of
MCMC methods. 
Deep learning method has been widely applied to solve the problem with multiscale features \cite{wang2020reduced, zhang2020learning, wang2020deep}.
In this work, we will recourse to the reinforcement learning approach and propose to use the reinforcement learning to accelerate the MCMC sampling.

Reinforcement learning (RL) has been studied since the last century. With the development of deep learning, deep reinforcement learning has been applied to solve various problems. 
Model-free RL algorithm is one of the two most common RL algorithms. It usually includes Q-learning and policy iteration methods (actor critic) \cite{kulkarni2016hierarchical, haarnoja2017reinforcement, lillicrap2015continuous,wang2016dueling, schulman2017proximal,mnih2015human,mnih2015human,
schaul2015prioritized, wang2016dueling, mnih2016asynchronous, haarnoja2018soft,haarnoja2018soft, fujimoto2019off, fujimoto2018addressing, Schulmanetal_ICLR2016, russell2016artificial, Williams:92, Sutton1998}. 
In this work, we are going to study the combination of the policy iteration method with the MCMC algorithms.
The target of the policy iteration RL is to learn an acceptable policy such that the expected future reward is maximized. 
Most of the models have a single agent that is in charge of predicting the behavior; however, there are problems which require the interplay (cooperation, competition) among multiple agents. Many multi-agent algorithms then have been proposed \cite{iqbal2019actor, zhao2019multi, lowe2017multi, foerster2017counterfactual, foerster2017stabilising}. One of the essential principles of multi-agent design is to centralized the critics but decentralized the actors. This means that the actors make decisions using local information while the critic is centralized learnt using global information. 

In this work, we proposed an RL alternative to accelerate the MCMC. The main idea is to use the RL policy as the proposal generator in the MCMC process. More precisely, 
the RL agent is learning a distribution of actions, and we can make a new proposal by sampling an action from this distribution; that is, the action sampled will modify the current state and we hence obtain the new proposal. The MCMC then comes and either rejects or accepts this proposal as standard. Since the proposal generated may be rejected by MCMC which involves heavy computation, it is natural to use the off-policy algorithm; that is, instead of generating an entire trajectory which is used to learn the critic in the RL, we only run the policy for one time and get a state action state transaction. This saves much time, however, brings in the exploration issue. We hence proposed to apply the $\epsilon$-greedy strategy to overcome this issue. To be more specific, we use the RL policy as the proposal with a certain probability $p$ and use the random walk, which is guaranteed to convergence with probability $1-p$.

We will verify our method by solving a time-dependent flow inverse problem.
Inversely locating the high permeability channels given the measurements is a challenging problem. Our goal is to identify the heterogeneous permeability field such that the corresponding solution is closed to the measurement (observation). 
To be more specific,
the underlying model of the inversion problem is a parabolic equation as follows:
\begin{eqnarray}
\begin{split}
    u_t-\nabla\cdot(\kappa \nabla u) &= f(x,t)&\quad \text{ in } D\times[0, T],\\
    u(x, 0) &= g(x)& \quad x \in D, \\
    \frac{\partial u}{\partial \bold{n}} &= h(x,t)& \quad t\in[0, T] \text{ and } x\in\partial D,
\end{split}
\end{eqnarray}
where $D \subset \mathbb{R}^d$ (with $d  \in \{ 2,3\}$);  $\partial D$ denotes the boundary of the domain and $\bold{n}$ is the unit outward normal vector; 
$\kappa(x)$ is a high-contrast permeability field and contains some channelized features. 
The measurement is the solution of the model at given region and the target is to find the permeability $\kappa(x)$ such that $\|F-\mathcal{F}(\kappa(x)) \|$ is small, where $F$ is the observation and $\mathcal{F}(\cdot)$ is the forward solver by which we use to calculate the solution of the input permeability field.

We achieve our goal by sampling a sequence of permeability fields which are characterized as the state in the reinforcement learning framework.
The sampling is supervised by the RL agent, and the new sampling will be either rejected or accepted by the MCMC algorithm.
We solve a challenging problem with multi-channels and
use a multi-agents RL algorithm. The basic idea is to use agents to capture the channels. Each agent is decentralized to learn a policy which will locate one channel. All agents work cooperatively, and the critic is centralized learnt. The more concrete setting will be presented in Section \ref{experiments}. 
To summarize our contribution, we proposed an algorithm which uses multi-agent RL to accelerate the MCMC sampling. We verify our method by solving two challenging inverse problems with the multiscale property. 

The rest of the work is organized as follow. In Section \ref{background}, we will introduce the preliminaries of the works. These include the generalized multiscale finite element method, reinforcement learning and the probabilistic MCMC formulation. Our proposed method is presented in Section \ref{preoposed}. We conduct several numerical experiments in Section \ref{experiments} to verify our idea.

\section{Background}
\label{background}
\subsection{Generalized Multiscale Finite Element Methods (GMsFEM)}
\label{gmsfem}
We are going to solve an inverse problem with multiscale property. To speed up the rejection process of the multilevel MCMC, we need to build coarse solvers. In this work, we are going to use the GMsFEM as a coarse solver. Consider the following time-dependent problem:
\begin{align*}
    Mu_t+Lu = f,
\end{align*}
defined in some region $\mathcal{I}\times D$. Here $M$ is a positive definite linear operator, $f$ is a random source function with sufficient regularity. 
For our model problem, $\mathcal{I} = [0, T]$, $D  = [0,1]^d$ (with $d \in \{2,3 \}$), $M = 1$ and $Lu := -\nabla\cdot(\kappa \nabla u)$. We remark that the permeability $\kappa \in L^\infty(D)$ contains some multiscale features. 

\begin{figure}[H]
\centering
\includegraphics[scale = 0.2]{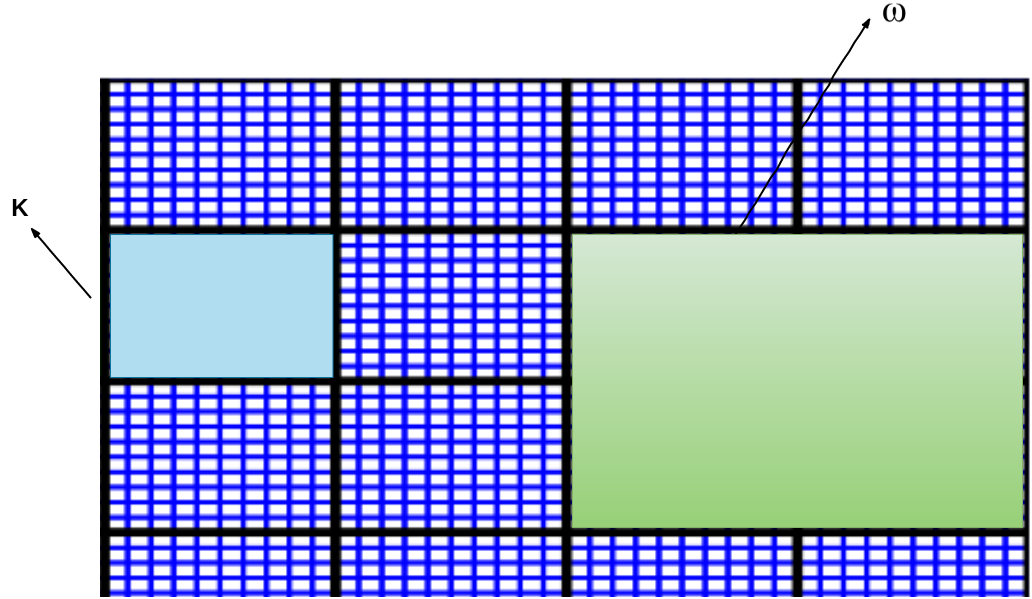}
\caption{Domain Partition $\mathcal{T}^H$.}
\label{grid}
\end{figure}

Let the spatial domain $D$ be partitioned by a coarse grid $\mathcal{T}^H$; this does not necessarily resolve the multiscale features. Let us denote $K$ as one coarse cell in $\mathcal{T}^H$. We refine $K$ to obtain the fine grid partition $\mathcal{T}^h$ (blue box in Figure~\ref{grid}). We assume the fine grid is a conforming refinement of the coarse grid. 
See Figure~\ref{grid} for~details. 

For the $i$-th coarse grid node, let $\omega_i$ be the set of all coarse elements having the vertex $i$ (green region in Figure~\ref{grid}).
We will solve local problem in each coarse neighborhood to obtain set of multiscale basis functions $\{ \phi_i^{\omega_i}\}$
and seek solution in the form:
\begin{align}
    u = \sum_i\sum_j c_{ij}\phi_j^{\omega_i},
\end{align}
where $\phi_j^{\omega_i}$ is the offline basis function in the $i$-th coarse neighborhood $\omega_i$ and $j$ denotes the $j$-th basis function. 
Before we construct the offline basis, we first need to derive the snapshot~basis.

\subsubsection{Snapshot~Space}
There are several ways to construct the snapshot space; we will use the harmonic extension of the fine grid functions defined on the boundary of $\omega_i$.  Let us denote $\delta_l^h(x)$ as fine grid delta function, which is defined as $\delta_l^h(x_k) = \delta_{l,k}$ for $x_k\in J_h(\omega_i)$ where $J_h(\omega_i)$ denotes the boundary nodes of $\omega_i$. The~snapshot function $\psi_l^{\omega_i}$ is then calculated by solving local problem in $\omega_i$:
\begin{align}
    L(\psi_l^{\omega_i}) = 0
\end{align}
subject to the boundary condition $\psi_l^{\omega_i} = \delta_l^h(x)$. The~snapshot space $V_{snap}^{\omega_i}$ is then constructed as the span of all snapshot~functions.

\subsubsection{Offline~Spaces}
The offline space $V_{\off}^{\omega_i}$ is derived from the snapshot space and is used for computing the solution of the problem. We need to solve for a spectral problem and this can be summarized as finding $\lambda$ and $v\in V_{\snap}^{\omega_i}$ such that:
\begin{align}
    a_{\omega_i}(v,w) = \lambda s_{\omega_i}(v,w), \forall w\in V_{\snap}^{\omega_i},
    \label{eig}
\end{align}
where $a_{\omega_i}$ is symmetric non-negative definite bilinear form and $s_{\omega_i}$ is symmetric positive definite bilinear form. By~convergence analysis, they are given by
\begin{align}
    &a_{\omega_i}(v,w) = \int_{\omega_i}\kappa \nabla v\cdot \nabla w,\\
    &s_{\omega_i}(v,w) = \int_{\omega_i} \tilde{\kappa} v\cdot w.
\end{align}

In the above definition of $s_{\omega_i}$, the~function $\tilde{\kappa} = \kappa \sum |\nabla \chi_j|^2$
where $\{ \chi_j\}$ is a set of partition of unity functions corresponding to the coarse grid partition of the domain $D$ 
and the summation is taken over all the functions in this set. 
The offline space is then constructed by choosing the smallest $L_i$ eigenvalues and we can form the space by the linear combination of snapshot basis using corresponding~eigenvectors:
\begin{align}
    \phi_k^{\omega_i} = \sum_{j = 1}^{L_i} \Psi_{kj}^{\omega_i}\psi_j^{\omega_i},
\end{align}
where $\Psi_{kj}^{\omega_i}$ is the $j$-th element of the $k$-th eigenvector and $L_i$ is the number of snapshot basis. 

It should be noted that, to construct the conforming basis, the selected eigenfunctions are multiplied by the partition of unity functions.
The coarse solvers $\mathcal{F}_l$ are defined using different number of offline basis $\Phi_k^{\omega_i}$. To be more specific, we will form space $V_{\off}$ as the collection of all local offline basis and then seek a solution $u_{\off}\in V_{\off}$ such that:
\begin{align}
    \left (\frac{d u_{\off}}{dt}, v \right )+a(u_{\off},v) =(f,v) \quad \forall v\in V_{\off}
    \label{linear}
\end{align}
where $a(u,v) := \int_D \kappa \nabla u \cdot \nabla v$ and $(u, v) := \int_{D}uv$. The choices of the time discretization are flexible and for example, we can use the backward Euler scheme.

\subsection{Actor Critic Reinforcement Learning}
We consider the multi-agent extension of the Markov decision processes (MDPs).
A Markov game with $N$ agents is defined by a set of state $\mathcal{S}$ and a set of actions $\mathcal{A}_1, ..., \mathcal{A}_N$ and a set of observations $\mathcal{O}_1, ..., \mathcal{O}_N$ for each agent $i$. To choose an action $a_i$, each agent will follow the policy $\pi_{\theta_i}(a_i|s_i): \mathcal{O}_i\times\mathcal{A}_i\rightarrow [0, 1]$, which is a distribution of the actions given current observation $s_i$; the policy $\pi_{\theta_i}(.)$ depends on the parameter $\theta_i$ and can be formulated as a network. The agents will then move to the next state $s' = \{s'_1, ..., s'_N\}$ by sampling from the transition probability distribution $T (s'|s, a) \in [0, 1]$, where $s$ is the set of current observations and $a = \{a_1, ..., a_N\}$ is the set of actions of each individual agent. Each agent $i$ obtains a reward $r_i(s_i, a_i): \mathcal{O}_i\times \mathcal{A}_i\rightarrow \mathbb{R}$ given the current observation $s_i$ and the action $a_i$ determined by the policy; the reward of all agents can be denoted as $r(s, a): \mathcal{S}\times \mathcal{A}\rightarrow \mathbb{R}$ where $\mathcal{A} = \mathcal{A}_1\times...\times\mathcal{A}_N$ is the set of all actions.

We denote the trajectory of length $T$ formed in this Markov process as $\tau: \mathcal{S}\times \mathcal{A}\times \mathcal{S}\times \mathcal{A}...\mathcal{S}\times \mathcal{A}$, the associated probability will then be denoted as $p_{\theta}(\tau)$. The target of the reinforcement learning is to maximize the expected return: 
$$J(\theta) = \mathbb{E}_{\tau}[R],$$ 
where $R$ is the reward of the trajectory and $\theta = \{\theta_i, ..., \theta_N\}$ are the parameters associated with the policy of each agent;
to be more specific,

\begin{align*}
    \max_{\theta}J(\theta) = \max_{\theta} E_{\tau\sim p_{\theta}(\tau)}
\left[
\sum_{t} r(s^t, a^t)
\right ],
\end{align*}
where $(s^t, a^t)$ is the state-action pair at step $t$ in the trajectory $\tau$.
This optimization process can be implemented by moving in the direction $\nabla J(\theta)$ which is called policy gradient. 

There are several approaches to estimate the expected return.
The most simple choice is the REINFORCE algorithm proposed by Sutton \cite{Williams:92}, that is, 
$R^t = \sum_{j = t}^T \gamma^{j-t}r^j$, where $\gamma$ is the discount factor which determines how much the policy favors the immediate reward over the long-term gain; to be more specific,
\begin{align}
\nabla_{\theta} J(\theta)=
\mathbb{E}_{\tau\sim p_{\theta}(\tau)}
\left[
\left(
\sum_{t = 1}^T \nabla_{\theta}\log\pi_{\theta}(a^t|s^t)
\right )
\left(
\sum_{t' = t}^T \gamma^{t'-t}r(s^{t'}, a^{t'})
\right)
\right ],
\end{align}
The REINFORCE algorithm usually leads to high variance and this issue can be ameliorated by approximating the expected return. One choice is to approximate the state value function $Q(s, a): \mathcal{S}\times\mathcal{A}\rightarrow\mathbb{R}$ which is defined recursively \cite{Sutton1998} as:
\begin{align}
    Q(s, a) = \mathbb{E}_{s'}[ r(s, a)+\gamma \mathbb{E}_{a'\sim \pi_\theta}(Q(s', a'))].
\end{align}
The policy gradient will then become:
\begin{align}
\nabla_{\theta} J(\theta)=
\mathbb{E}_{\tau\sim p_{\theta}(\tau)}
\left[
\left(
\sum_{t = 1}^T \nabla_{\theta}\log\pi_{\theta}(a^t|s^t)
\right )
Q(s^t, a^t)
\right ].
\end{align}
By varying the advantage function $A$ (state value function $Q$ and REINFORCE are two special cases), many practical algorithms have been proposed \cite{Schulmanetal_ICLR2016}. In this work, we will consider the advantage function defined as:
\begin{align}
    A(s, a) = Q(s, a)-V(s),
\end{align}
where $V(s): \mathcal{S}\rightarrow \mathbb{R}$ is the state value function which is the expectation of $Q$,
\begin{align}
    V(s) = \mathbb{E}_{a\sim\pi} Q(s, a). 
\end{align}
The approximation of the advantage function have been thoroughly studied in \cite{Schulmanetal_ICLR2016}. In this work, we will use:
\begin{align}
    r(s, a)+\gamma V_{\omega}(s')-V_{\omega}(s),
\end{align}
where $V_{\omega}$ is the approximation of the value function and can be formulated as a network with parameter $\omega$. This approximation is a special case of the $\gamma-$just (unbiased) operator \cite{Schulmanetal_ICLR2016} when $V_{\omega} = V$ and typically has lower variance. We now introduce the basic actor-critic algorithm here:
\begin{enumerate}
    \item Generate trajectories by policy $\pi_{\theta}$.
    \item Get the updated advantage function $A_{\omega}$ by learning $V_{\omega}(.)$ with the trajectories.
    \item Update the policy $\pi_{\theta}$ by optimize the objective $J(\theta)$:
    $$
    \nabla_{\theta} J(\theta)=
\mathbb{E}_{\tau\sim p_{\theta}(\tau)}
\left[
\left(
\sum_{t = 1}^T \nabla_{\theta}\log\pi_{\theta}(a^t|s^t)
\right )
A(s^t, a^t)
\right ].
    $$
\end{enumerate}
It should be noted that we use the off-policy algorithm in practice. That is, in step 1 above, we run the policy for one time and get a state action state transaction. This transaction will be saved into reply buffer and we train the value function in step 2 by sampling some transactions from the reply buffer. The off-policy algorithm is fast but may result in the local convergence. We will discuss this issue in detail in the numerical example sections.
In this work, we use the actor-critic (AC) algorithm incorporated with the MCMC algorithm. The proposed algorithm will be detailed in later sections.

\subsection{Probabilistic MCMC}
\label{prob_formulation}
This problem can be summarized as sampling the permeability field which has high contrast channels given pressure data with some measurement error. Denote $P(\kappa|F)$ as the conditional probability of the permeability field $\kappa$ given the observation $F$, we then will sample permeability from the $P(\kappa|F)$. By the Baysian' formula, it follows that:
$$
p(\kappa|F) \propto P(F|\kappa)\cdot P(\kappa),
$$
where $P(\kappa)$ is prior, $P(F|\kappa)$ is the likelihood and $\Pi(\kappa) = p(\kappa|F)$ is the posterior.
There are some errors associated with the likelihood function. The first to mention is the error of the observation of $F$. The second error comes from the forward solver denoted as $\mathcal{F}$ which solves the model problem given the sample permeability $\kappa$. We will assume the total errors follows the normal distribution with standard deviation $\sigma_f$; this gives us that,
\begin{align}
    P(F|\kappa) \propto \exp \left \{ {-\frac{\|F-\mathcal{F}(\kappa)\|_2^2}{\sigma_f^2}} \right \}. 
\end{align}
Sampling from $\Pi(\kappa)$ can be accomplished by the MCMC methods. In this work, we will build the algorithm basing on the Metropolis-Hasting algorithm \cite{hastings1970monte}.
It should be noted that in the computation of the posterior $\Pi(\kappa)$, intense computation will be involved in evaluating the forward solver $\mathcal{F}$. To improve the efficiency of computing the acceptance rate and give fast rejection, we adopted the multi-level MCMC algorithm basing on the GMsFEM.

We use GMsFEM coarse solvers $\mathcal{F}_1, ..., \mathcal{F}_L$ defined in Section \ref{gmsfem};
then the corresponding posterior will be denoted as:
\begin{align}
    \Pi_l \propto \exp \left \{ {-\frac{\|F-\mathcal{F}_l(\kappa)\|_2^2}{\sigma_f^2}} \right \}P(\kappa).
    \label{Pi}
\end{align}
We denote $q(y|x)$ the proposal generator in which $x$ is the current sampling and $y$ is the proposal. The multi-level MCMC algorithm is detailed in Appendix \ref{MCMC}.

\section{Proposed Approach}
\label{preoposed}
The traditional MCMC algorithm can be accelerated by multi-level MCMC since the multilevel MCMC uses coarse solver which gives faster rejection computation; however MCMC can be further improved by modifying the proposal generator. We hence proposed the multi-agent reinforcement approach.

\subsection{Explanations of the Algorithm}
Each channel in the target permeability field can be parametrized and taken as observations ($s_i$) in the RL algorithm; and each agent is in charge of chasing for one target channel of quadrilateral shape. Actions ($\mathcal{A}_i$) are defined so that each rectangle can be reshaped or moved; and action distribution of each agent is determined by the policy $\pi_{\theta_i}(.|s_i)$.
Detailed setup can be seen in Section \ref{setup}.

To be more clear, each agent is like a robot; at the beginning of the simulation, each robot stands at some position in the field, then they can transform and move individually so that the output of the forward solver at the end of simulation is closed to the real observation.
The pseudo algorithm has 3 iterative steps and please check Appendix \ref{rl_mcmc} for the details of the algorithm:

\begin{enumerate}
    \item Each RL agent return a new local proposal for its own channel according to its own policy; make a new state by combining all proposals.
    \item Apply standard multilevel MCMC to accept or reject the new state. If the new state is rejected, go back to step 1; otherwise proceed to the step 3.
    \item  Update the RL agents (policy and critic) using the past trajectory offline and then go back to step 1.
\end{enumerate}
The first step above is the sampling process in MCMC algorithm and we replace the random proposal $q(y|x)$ by the reinforce. More precisely,
the sampling (for a single agent) is achieved by following three steps below:
\begin{enumerate}
    \item Given current state $s_i$.
    \item Sample an action $a_i$ from the policy $\pi_{\theta_i}(.|s)$.
    \item Execute action $a_i$ to get the next state $s_i'$.
\end{enumerate}
That is,
$
q(s'_i|s_i) = \pi_{\theta_i}(a_i|s_i),
$
such that $s'_i$ is acquired by taking action $a_i$ at state $s_i$. 

\subsection{Discussions of the Algorithm}
There are some key notes of the algorithm.
\begin{enumerate}
    \item Due to the evaluation of the forward solvers in the MCMC step, it is not realistic to run the policy multiple times and get a long enough trajectory to update the RL agents (step 3).
We hence apply the off-policy algorithm as discussed before. This may result in the local convergence since the RL agent may never explore some regions in the solution space and we are learning the value function greedily.
This exploitation (or the under exploration) issue will result in the early stopping in our problem. More precisely, proposals with large probabilities suggested by the RL policy will be very likely rejected by MCMC; meanwhile, the low probabilities actions are easily to be accepted by MCMC but are hard to be sampled. Consequently, the sampling process becomes slow and we call this phenomenon ``early stopping".
We will demonstrate this phenomenon in details later on the numerical example sections; but to
solve this issue, we consider the $\epsilon$-greedy strategy.
That is, we use the RL policy as the proposal with a certain probability $p$ and use the random walk proposal which is guaranteed to convergence with probability $1-p$.
\item We use the the decentralized actor and centralized critic principle. To be more specific, the policy $\pi_{\theta_i}(.|s_i)$ of each agent $i$ predict the action distribution depending only on the local observation $s_i$. Once agent $i$ moves to new observation $s_i'$ and obtains the new state $\hat{s}' = \{s_1',...,s_i', s_{i+1}, s_N\}$, we can evaluate the reward $r$ using the global information $\hat{s}'$; the critic value function $V_i(.|\omega_i)$ will then be updated accordingly. Intuitively, the centralized critic is able to enhance the cooperation among all agents. The expected value reward ($V_i(|\omega_i)$) for one observation-action pair (of one agent) relies on the global state information; hence the agent will not make decision greedily only improving its own reward, instead, individual move will be beneficial to the global reward. In our problem, it is natural to evaluate the forward solver using global information, that is, we need all channels in order to compute the reward; hence the centralized critic should be a good choice.

\item (Intuition of the idea) Instead of moving to a state greedily, the RL algorithm is supposed to move to a state which has better expected reward to go. This is accomplished by the critic learning in the RL iteration. Hence, the proposal generated by the RL agent should give faster convergence than the random sampling.
\end{enumerate}

\section{Numerical Examples}
\label{experiments}
In this section, we are going to demonstrate 3 sets of the numerical experiments. The first experiment (in Sections \ref{exp1_1} - \ref{exp1_4}) and the third one (in Section \ref{exp3}) are aimed at showing the efficiency of the proposed method; while the second experiment in Section \ref{early} is conducted to explain the early stopping of the RL method. The RL setup is presented in Section \ref{setup}.

\subsection{Reinforcement Learning Setup}
\label{setup}
In this section, we will model our problem under the reinforcement learning framework. There are 3 basic elements that we need to define to model our problem: $\mathcal{S}$, $\{\mathcal{O}\}_{i = 1}^N$, $\{\mathcal{A}\}_{i = 1}^N$ and the reward $r(s, a)$.

The target of the inversion problem is to find the high contrast channel such that the observation loss is minimized. In the framework of the RL, each agent is in charge of one channel and moves to the target channels step by step.
To simplify the setting of the problem, we assume all channels have quadrilateral shape and we hence characterize a channel by a tuple of 4 parameters $(x,y,w,d)$, where $(x,y)$ is the coordinate of the lower left vertex of the channel while $(w, d)$ stands for the width and height.
The observation of each agent is then defined as:
\begin{align}
    \mathcal{O}_i =\{(x,y,w,d)\}, i = 1,..., N
\end{align}
and the state is consisted of all observations and is then defined as:
\begin{align}
    \mathcal{S} = \{(x_1,y_1,w_1,d_1),..., (x_N,y_N,w_N,d_N)\},
\end{align}
where $N$ is the number of channels. Since we are using the decentralized actor, each agent will modify its prediction basing on the local information; we hence define the 8 actions for each agent. Also, we assume the transition is deterministic and then the state and action can be summarized as follow: 
\begin{align*}
&T\big( (x-h, y, w, d)|(x, y, w, d), a \big) = 1, a = \text{shift to the left} \\
&T\big((x+h, y, w, d)|(x, y, w, d), a\big) = 1, a = \text{shift to the right} \\
&T\big((x, y-h, w, d)|(x, y, w, d), a\big) = 1, a = \text{shift downwards} \\
&T\big((x, y+h, w, d)|(x, y, w, d), a\big) = 1, a = \text{shift upwards} \\
&T\big((x, y, w-h, d)|(x, y, w, d), a\big) = 1, a = \text{squeeze horizontally} \\
&T\big((x, y, w+h, d)|(x, y, w, d), a\big) = 1, a = \text{stretch  horizontally} \\
&T\big((x, y, w, d-h)|(x, y, w, d), a\big) = 1, a = \text{squeeze vertically} \\
&T\big((x, y, w, d+h)|(x, y, w, d), a\big) = 1, a = \text{stretch vertically},
\end{align*}
where $h$ is the size of one fine element. We assume the deterministic transition, hence define the reward as 
$$
r(s, a) = r(s, s') = \|\mathcal{F}-\mathcal{F}_L(s)\|-c_1 \|\mathcal{F}-\mathcal{F}_L(s')\|+c_2,
$$
where $(s, s')$ are the current and proceeding states respectively; $c_1$ and $c_2$ are two hyper-parameters to set; $\mathcal{F}_L$ is the coarse scale forward solver and $\mathcal{F}$ is the observation. Throughout all the experiments, the observation is the pressure at wells.

We random choose the starting position for each channel and run the algorithm. At each step, we compute  $\|\mathcal{F}-\mathcal{F}_L(s)\|$. We will demonstrate the convergence of the multilevel MCMC (denoted as MCMC), MCMC improved by RL (RLMCMC) and an $\epsilon$-greedy update version of the RLMCMC algorithm (eRLMCMC).

\subsection{The First Experiment Setup}
\label{exp1_1}
\begin{figure}[H]
\centering
\mbox{
\includegraphics[scale = 0.4]{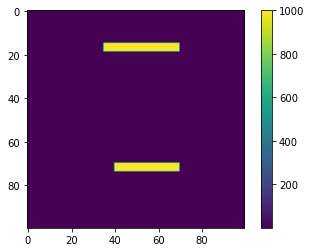}
\includegraphics[scale = 0.4]{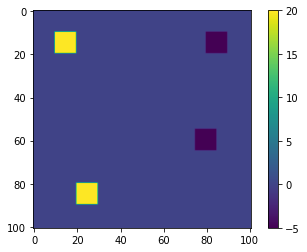}
\includegraphics[scale = 0.4]{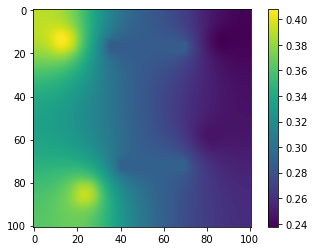}
}
\caption{Left: the target permeability field. The permeability of the field at the channels is equal to 1000 and is equal to 1 otherwise. Middle: source of the system. Right: corresponding solution.}
\label{system}
\end{figure}

The target permeability field with two high contrast channels are shown in Figure \ref{system}. The source with $2$ injection wells and $2$ production wells is defined as follows (see Figure \ref{system} for the illustration):
\begin{eqnarray}
f(x) = \left \{
\begin{array}{ll}
20,  x \in [0.1, 0.2] \times [0.1, 0.2], \\
\\
-5,  x \in [0.8, 0.9] \times [0.1, 0.2], \\
\\
20, x \in [0.2, 0.3] \times [0.8, 0.9], \\
\\
-5, x \in [0.75, 0.85] \times [0.55, 0.65].
\end{array}
\right .
\end{eqnarray}
The solution of the system is shown in Figure \ref{system} and we obtain the solution using the fine mesh ($h = 1/100$). It should be noted that we only use the data at the given wells as the observation data. 

\subsection{Multilevel MCMC}
\label{exp1_2}
\begin{figure}[ht]
\centering
\includegraphics[width = 2.5in]{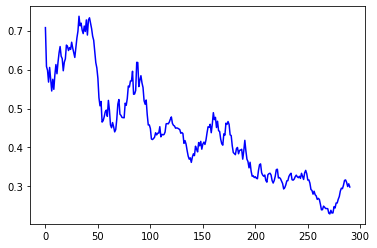}
\caption{Convergence of the Multilevel MCMC. The $y$-axis is: $\|\mathcal{F}-\mathcal{F}_L(s)\|$ and the $x$-axis is the steps of the sampling.} 
\label{fig_mcmc}
\end{figure}

We will first demonstrate that the conventional MCMC algorithm. It should be noted that we do not use prior in the algorithm, that is, $P(\kappa) = 1$ in \eqref{Pi}. Also, we use the uniform proposal, i.e., each action has the equal probability to be sampled.
Please see Figure \ref{fig_mcmc} for the convergence of the MCMC method.
In Figure \ref{fig_mcmc}, 
the $x$-axis is the sampling steps and please note that, only the accepted samplings are recorded; while the $y$-axis represents the difference between the measurement and the solution.
Our large scale experiments show that the convergence speed of the traditional method is slow when compared to the proposed method which is shown in the next section.

\subsection{RL accelerated MCMC (RLMCMC)}
\label{exp1_3}
\begin{figure}[H]
\centering
\mbox{
\includegraphics[scale = 0.4]{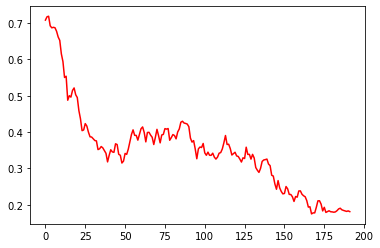}
\includegraphics[scale = 0.4]{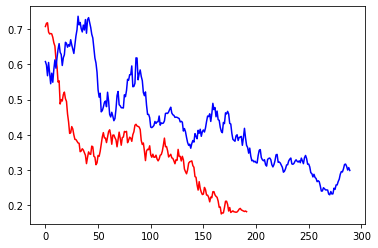}
}
\caption{Left: the convergence of the the RL-MCMC method. Right: the comparison of the MCMC method and the proposed method. RED: Proposed; BLUE: MCMC.}
\label{first_comp}
\end{figure}

In this experiment, we accelerate the conventional MCMC by the reinforcement learning. To be more specific, the proposal is replaced by the RL agent policy. All the other hyper-parameters will be kept the same as the previous experiment. Please check Figure \ref{first_comp} (Left) for the convergence of the RL-MLMCMC; the comparison of two method can be seen in Figure \ref{first_comp} (Right).

The CPU time for the computation can be seen in Table \ref{time1}.
We can see the faster convergence of the RLMCMC method; however, we also observe the early stopping of the RLMCMC, that is, almost all new proposals proposed by RL agents will be rejected by the MCMC algorithm. In this experiment, this phenomenon happens at around $100$ steps. This will slow down the convergence of the method. We will explain the reason of this in the next section and give an update of the method in the next subsection.

\subsection{Early Stopping}
\label{early}
We observed the early stopping phenomenon of the RLMCMC; that is, almost all new proposals suggested by the actors are rejected by the MCMC algorithm.
The good news is, our large scale experiments show that the stopping predictions have satisfying results and this stopping can be controlled by the adjusting the learning rate. Please check the Figure \ref{first_comp} (Left).

The direct reason of the early stopping is that the probabilities for two actions which are in a pair (e.g., move to the left and move the right) have multiple scales as RL learning goes. By the formula of the calculating the acceptance rate \eqref{mlmcmc}, the acceptance rate becomes extremely low and hence the new proposal which is sampled with large probability suggested by the RL agent is rejected. This causes the slow sampling process.

The fundamental reason of the multi-scales probabilities is the RL exploration problem. We use the off-policy RL, that is, the value function is updated from samples in the reply buffer.
The benefits of the off-policy strategy is: we do not need to run the policy for multiple times and get a trajectory, which is super time consuming.
That means if the agent never explores a region, it has no information about that region and hence makes decision only depending on the region it has explored. To explain this issue better, let us consider a simpler model.

\begin{figure}[H]
\centering
\mbox{
\includegraphics[scale = 0.4]{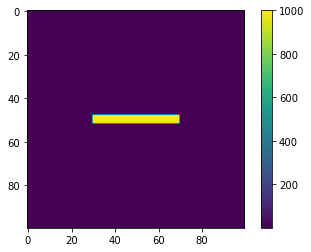}
\includegraphics[scale = 0.4]{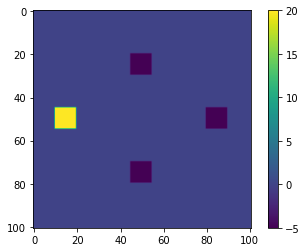}
\includegraphics[scale = 0.4]{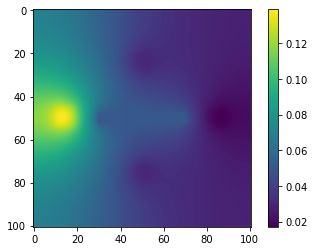}
}
\caption{One channel example. Left: the target permeability field. The permeability of the field at the channels is equal to 1000 and is equal to 1 otherwise. Middle: source of the system. Right: corresponding solution.}
\label{system2}
\end{figure}

The target permeability has only one channel (see Figure \ref{system2} (Left)); and we have 1 injection well and 3 production wells (see Figure \ref{system2} (Middle)). That is, the source is defined as:
\begin{eqnarray}
f(x) = \left \{
\begin{array}{ll}
-5,  x \in [0.45, 0.55] \times [0.2, 0.3], \\
\\
-5,  x \in [0.45, 0.55] \times [0.7, 0.8], \\
\\
20, x \in [0.1, 0.2] \times [0.45, 0.55], \\
\\
-5, x \in [0.8, 0.9] \times [0.45, 0.55]. 
\end{array}
\right .
\end{eqnarray}
In this example, 
we assume we know $(y, w, d)$ and only $x$ is unknown, that means the agent only needs to move to the right or left. 
We use one agent. State (observations) will be formulated same as before; however, the action space contains two actions: move to the left and move to the right. To better demonstrate the probability evolution and avoid the early stopping caused by the MCMC rejection, we use only RL to make the sampling and there is no MCMC involved.

\begin{figure}[H]
\centering
\mbox{
\includegraphics[scale = 0.4]{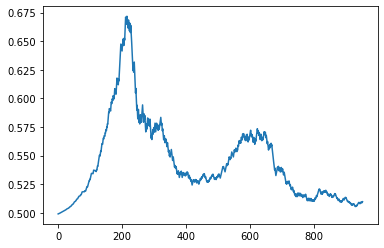}
\includegraphics[scale = 0.4]{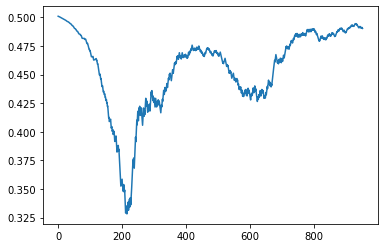}
}
\caption{Left: Probability of going left. Right: Probability of going right. In both graphs, the y axis is the probability and x axis is the training step.  }
\label{leftright}
\end{figure}

The guess starting position will be to the right of the target place. This means that the agent only has information of moving to the left; hence the probability of moving to the left increases drastically before the agent passes over the target position; the multiscale in probabilities happens then. In the formula of the acceptance rate,
$q(s^m|c)$ is the probability of moving to the right which is small compared to $q(c|s^m)$ which is moving to the left. This results in the low acceptance rate.
Please check Figure \ref{leftright} for the illustration of the evolution of the probability distribution. 

Our experiments show that the early stopping prediction has been good enough when compared to the traditional MCMC algorithm; however, this brings the under-exploration issue to the RL algorithm. To solve this issue, we employ the $\epsilon-greedy$ method to upgrade the RLMCMC algorithm.

\subsection{$\epsilon-greedy$ RL-MCMC (eRLMCMC)}
\label{exp1_4}
To solve the early stopping issue the RL-MCMC method, we proposed the $\epsilon-greedy$ RL-MCMC (eRLMCMC) method. The idea is to apply the $\epsilon-greedy$ strategy and mix the RL-MCMC and MCMC method. To be more specific, in each step, we use RL policy as proposal with the probability $\epsilon$ and use uniform policy with probability $1-\epsilon$. 

\begin{figure}[H]
\centering
\mbox{
\includegraphics[scale = 0.4]{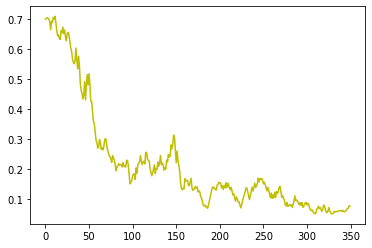}
\includegraphics[scale = 0.4]{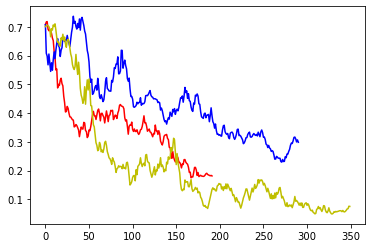}
}
\caption{Left: the convergence of the the eRLMCMC method. Right, the comparison of all 3 methods. RED: RLMCMC, BLUE: MCMC, YELLOW: eRLMCMC}
\label{2comp}
\end{figure}

Please check Figure \ref{2comp} (Left) for the convergence of eRLMCMC; and the computation time is shown in Table \ref{time1}.

\begin{table}[H]
\centering
\begin{tabular}{||c c c||} 
\hline
Method & Number of Steps & Time \\ [0.5ex] 
\hline
MCMC & 289 & 8060.6s \\ [0.5ex]
\hline
RLMCMC & 191 & 7274.6s \\ [0.5ex]
\hline
eRLMCMC & 349 & 9680.4s \\ [0.5ex]
\hline
\end{tabular}
\caption{Computation time of the first example}
\label{time1}
\end{table}

In this table, we compared the computational time for all three methods. It should be noted that the average step time of the RLMCMC method (proposed method) is larger than the that of the traditional method. This is because of the computations needed in training the RL agent. However, we can observe that the RLMCMC method has shorter total computation time and better result (see Figure \ref{2comp} (Right)); this shows that the proposals suggested by the RL agent are meaningful and can lead to the convergence in a more efficient way. Similar results can be seen in the second examples Table \ref{time2}.
Our method hence works.

\subsection{Test on Diagonal Channels}
\label{exp3}
\begin{figure}[H]
\centering
\mbox{
\includegraphics[scale = 0.4]{graphs/source.png}
\includegraphics[scale = 0.4]{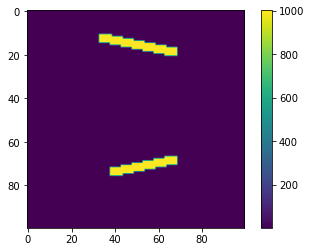}
\includegraphics[scale = 0.4]{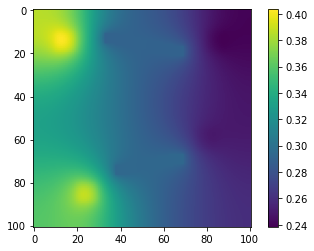}
}
\caption{Problem with the diagonal channels. Left: the target permeability field. The permeability of the field at the channels is equal to 1000 and is equal to 1 otherwise. Middle: source of the system. Right: corresponding solution.}
\label{system3}
\end{figure}

In the first set of experiments, we assume the underlying channels are horizontal or vertical; however there are cases in the real applications that the channels are diagonal. In this set of experiments, we will test more challenging example with diagonal channels; however, the state and action formulations are kept the same as before, that is, we still use the rectangles which are parallel with the domain boundaries as the predicting states (observations). We use the same source term but the channels are diagonal. Please check Figure \ref{system3} for the details.

\begin{figure}[H]
\centering
\mbox{
\includegraphics[scale = 0.4]{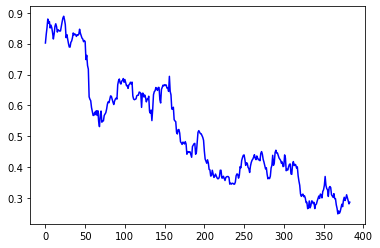}
\includegraphics[scale = 0.4]{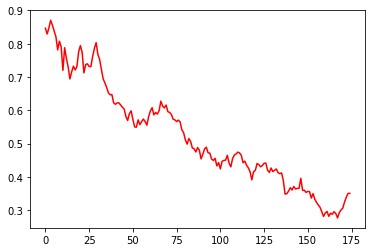}
\includegraphics[scale = 0.4]{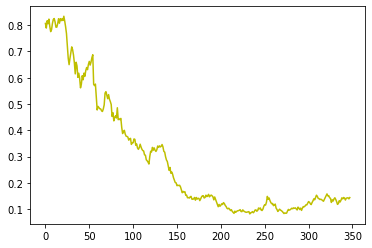}
}
\caption{Left: MCMC result. Middle: MCMC improved by RL (RLMCMC). Right: $\epsilon$-greedy strategy RLMCMC}
\label{diag_ind_results}
\end{figure}

\begin{figure}[H]
\centering
\mbox{
\includegraphics[scale = 0.4]{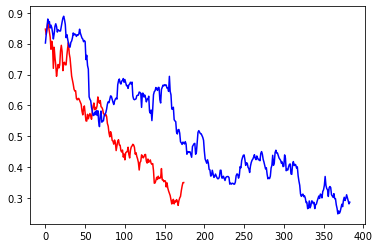}
\includegraphics[scale = 0.4]{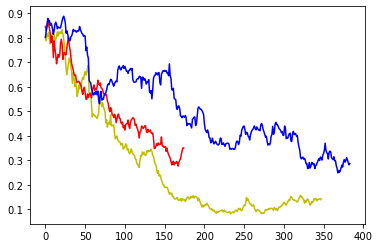}
}
\caption{Left: MCMC vs RLMCMC; blue curve is the MCMC and red curve is the RLMCMC. Right: MCMC vs RLMCMC vs eRLMCMC; blue curve is the MCMC, red curve is the RLMCMC and yellow is the eRLMCMC.}
\label{diag_comp}
\end{figure}

Same as before we will compute $\| \mathcal{F}-\mathcal{F}_L(s)\|$ and demonstrate the results of applying MCMC, RLMCMC and eRLMCMC. Please check Figure (\ref{diag_ind_results}) for the individual results of three methods. 
We also observe the early stopping of the RLMCMC method and hence use the $\epsilon$-greedy strategy to improve the RL algorithm. The comparison of three methods are shown in Figure \ref{diag_comp}. 
We can observe from Figure \ref{diag_comp} that the RLMCMC does improve the convergence speed of the MCMC method. The  $\epsilon$-greedy strategy further improves the result by extending the sampling process. The comparison of the computational time is shown is in Table \ref{time2}.
\begin{table}[H]
\centering
\begin{tabular}{||c c c||} 
\hline
Method & Number of Steps & Time \\ [0.5ex] 
\hline
MCMC & 383 & 8048s \\ [0.5ex]
\hline
RLMCMC & 174 & 6062.4.6s \\ [0.5ex]
\hline
eRLMCMC & 347 & 13472.4.4s \\ [0.5ex]
\hline
\end{tabular}
\caption{Computation time of the third example (Diagonal channels).} 
\label{time2}
\end{table}

From Table \ref{time2}, we can see that our methods still work.
One may notice the long total computation time of the eRLMCMC method. This happens because of the rejections. We apply the $\epsilon$-greedy strategy; however, setting $\epsilon$ is tricky. In this experiment, we set a relative large $\epsilon$. This means more rejections will be suggested by the MCMC and hence results in the long computation time; but this strategy indeed extends the trajectory and gives us a better convergence result. 

\newpage
\appendix

\section{Multi-level MCMC}
\label{MCMC}

\begin{algorithm}[H]
\SetAlgoLined
Set the total number of levels $L$\;
Given input $s^m$, sample a new proposal $c$ from the proposal generator $q(.|s^m)$\;
Compute the acceptance probability:
$$
\rho_0(c, s^m) = \min \left \{
1, \frac{\Pi_0(c) q(s^m|c) } { \Pi_0(s^m) q(c|s^m) }
\right \}\;
$$

Set $c_0 = c$ with the probability $\rho_0(c, s^m)$ and $c_0 = s^m$ with the probability $1-\rho_0(c, s^m)$\;
\For{$l = 1,...,L$}
{

Compute the acceptance probability by,
$$
\rho(c_{l-1}, s^m) = \min\left(
1, \frac{\Pi_{l-1}(s^m) \Pi_l(c_{l-1})  }{\Pi_{l-1}(c_{l-1})\Pi_l(s^m)}
\right)
$$
Set $c_l = c_{l-1}$ with the probability $\rho(c_{l-1}, s^m)$ and $c_l = s^m$ with the probability $1-\rho(c_{l-1}, s^m)$\;

}
Set $s^{m+1} = c_L$ and return $s^{m+1}$\;

\caption{MLMCMC}
\label{mlmcmc}
\end{algorithm}
It should be noted that once the proposal is rejected at some level, the following levels computation can be avoided since the accept probability is equal to $1$ and $c_l = s^m$ for all following $l$. This means that the proposal will be rejected by coarser solver which requires less computation time.

\newpage
\section{Multiagent RL Multilevel MCMC}
\label{rl_mcmc}

\begin{algorithm}[H]
\SetAlgoLined

Set the max length of the trajectory $T$\;
Initialize the critic net $V_l(.|\omega_l)$. the policy network $\pi_{\theta_l}(,|s)$ for all agent $l = 1, ..., C$\; 
Initialize the target net $\hat{V}_l(.|\omega'_l)$ by setting $\omega'_l\leftarrow \omega_l$\;
Initialize the reply buffer $\mathcal{D}$\;
Initialize the starting state $s^t = (s^t_1, ..., s^t_{C})$\;
\For{$t = 0,...,T$}
{
\For{$l = 0, ..., C$}
{
\While{$s^t_l == s^{t+1}_l$ }
{
Run $c = MLMCMC(s^t_l)$ with the proposal generator $q = \pi_{\theta_l}$\;
Set $s^{t+1}_l = c$\;
}

Random select $S$ samples $\{(\bold{s}_j, r_j, \bold{s}'_j)\}_{j = 1}^S$ from the reply buffer $\mathcal{D}$; the sample of agent $l$ will be denoted as $(s_{l, j}, s'_{l, j})$\;
Calculate the target for each sample by $V_{tar, l}(s_{l, j}|\omega_l') = r(\bold{s}_j, \bold{s}'_j)+\gamma \hat{V}_l(s'_{l, j}|\omega'_l)$ using the target network and the selected samples\;
Update the critic network $V_l(\omega_l)$ (over the batch) by optimizing,
$$
L_{V_l}(\omega_l) = \frac{1}{S}\sum_{j = 1}^S (V_l(s_{l, j}|\omega_l) - V_{tar, l}(s_{l, j}|\omega'_l))^2 \;
$$

Calculate the advancement $A_l(s_{l, j}, s'_{l, j}) = r(\bold{s}_j, \bold{s}'_j)+\gamma V_l(s'_{l, j}|\omega_l)-V_l(s_{l, j}|\omega_l)$ based on the critic network for all samples in the batch\;
Update the policy $\pi_{\theta_l}$ by optimizing
$$
L_{\pi_l} (\theta_l) = \frac{1}{S}\sum_{j = 1}^{S} (-A_l(s_{l, j}, s'_{l, j}) \log \pi_{\theta_l}(s'_{l, j}|s_{l, j})) \;
$$

Update the target network by:

$$
\omega_l'\leftarrow \epsilon\omega_l'+(1-\epsilon)\omega_l\;
$$
} 

Set $s^{t+1} = (s^{t+1}_1, ..., s^{t+1}_C)$ and store $(s^t, r(s^t, s^{t+1}), s^{t+1})$ in the buffer\;

} 

\caption{Replay Buffer A2C with MCMC}
\end{algorithm}
We have several remarks regarding the algorithm.
We need to call MLMCMC algorithm to generate samples; however, in the implementation of the MLMCMC algorithm, the proposal generation is completed in $2$ steps. First, we can sample an action from $\pi_{\theta_l}(.|s_l^t)$; and then we execute the action to get the proposal.

To initialize the reply buffer, we randomly generate samples $(s^t, r, s^{t+1})$; to get the reward of the transaction, we use the most coarse forward solver to evaluate the solution. This saves a lot of time in generating samples and the performance is not compromised. 

\bibliographystyle{abbrv}
\bibliography{references}
\end{document}